# A Linear-complexity Multi-biometric Forensic Document Analysis System, by Fusing the Stylome and Signature Modalities

Sayyed-Ali Hossayni[1,2,3*], Yousef Alizadeh-Q[1], Vahid Tavana[2], Seyed M. Hosseini Nejad [3], Mohammad-R Akbarzadeh-T [2,4], Esteve Del Acebo [3], Josep Lluis De la Rosa i Esteva [3], Enrico Grosso [5], Massimo Tistarelli [5], Przemysław Kudłacik [6]

[1] DMLab, School of Computer Engineering, Iran University of Science and Technology, Tehran, Iran.
[2] SCIIP center of excellence, Ferdowsi University of Mashhad, Mashhad, Iran.
[3] Agents Research Lab, TECNIO Centre EASY, University of Girona, Girona, Catalonia, Spain.
[4] Department of EECS, University of California, Berkeley, CA 94720-1776, USA
[5] Department of Sciences and Information Technology, University of Sassari, Sassari, Italy
[6] Department of Computer Science, University of Silesia, Bedzinska 39, 41-200 Sosnowiec, Poland
[*] hossayni@silver.udg.edu

**Abstract:** Forensic Document Analysis (FDA) addresses the problem of finding the authorship of a given document. Identification of the document writer via a number of its modalities (e.g. handwriting, signature, linguistic writing style (i.e. stylome), etc.) has been studied in the FDA state-of-the-art. But, no research is conducted on the fusion of stylome and signature modalities. In this paper, we propose such a bimodal FDA system (which has vast applications in judicial, police-related, and historical documents analysis) with a focus on time-complexity. The proposed bimodal system can be trained and tested with linear time complexity. For this purpose, we first revisit Multinomial Naïve Bayes (MNB), as the best state-of-the-art linear-complexity authorship attribution system and, then, prove its superior accuracy to the well-known linear-complexity classifiers in the state-of-the-art. Then, we propose a fuzzy version of MNB for being fused with a state-of-the-art well-known linear-complexity fuzzy signature recognition system. For the evaluation purposes, we construct a chimeric dataset, composed of signatures and textual contents of different letters. Despite its linear-complexity, the proposed multi-biometric system is proven to meaningfully improve its state-of-the-art unimodal counterparts, regarding the accuracy, F-Score, Detection Error Trade-off (DET), Cumulative Match Characteristics (CMC), and Match Score Histograms (MSH) evaluation metrics.

## 1. Introduction

Automatic systems are increasingly replacing traditional human-based identification methods, providing higher security and user convenience [1]. Biometric systems can be broadly classified into two categories: unimodal and multimodal [2]. The former category deals with a single source of information (e.g. only fingerprint or only face for recognition of a subject), which has to deal with different problems such as having noisy data, non-universality, or intra-class variations [3]. For overcoming the limitations of the former, the latter systems are proposed by integrating multiple sources of information. It makes the system to be less vulnerable to spoofing attacks, due to the difficulty of simultaneously spoofing multiple biometric traits. Also, due to the sufficient population coverage, multimodal Biometric systems are able to address the non-universality problem [3]. Information fusion of different modalities can be done in sensor-, feature-, score-, rank- or decision-level, each of which suitable for some applications. Depending on the to-be-fused modalities, one or some of the mentioned strategies can be adopted [4]. In addition to their more security, multi-biometric systems are also shown to be more efficient and reliable than unimodal systems.

One of the important applications of Biometrics is Forensic Document Analysis (FDA) [5]. FDA deals with different problems such as detection of the fraudulent documents either if they are modified or are totally imposter, checking the genuineness of some security features (e.g. signatures in cheques), identification of an unknown author/writer of a known document [6], investigating the existing methods for altering the documents, providing technical advising points for developing new security features for the official documents, analyzing the handwriting of the writers and the stylistic manners of the authors of documents and etc. [7]. Documents involve three main behavioral attributes; handwriting, signature, and linguistic style of writing (stylome). Although the state-of-the-art research studies consider the fusion of stylomes with other behavioral attributes [8], to the best of our knowledge, no research has been conducted on taking the synergetic advantages of fusing linguistic- and drawing- style of authors, whereas these two modalities exist together in many real-world

---

FDA applications such as the statement documents of defendants or witnesses ... in judicial or similarly in police-related paperwork, historical documents and etc.

In this paper, we take advantage of the synergetic effects of the mentioned FDA-related modalities by fusing them. We focus on the real-time applications. Therefore, for the signature modality, we utilize a linear-complexity signature recognition/verification algorithm in the state-of-the-art [9], and for the stylome modality, we investigate and introduce the most appropriate (in the same time accurate, both, fast and accurate) state-of-the-art method for authorship attribution. Considering the intrinsic difference of these modalities which inspires lack of correlation between them, fusing these modalities at sensor- and feature- level would not be meaningful. Thus, we propose their fusion at the score-level as the most commonly utilized level of fusion.

The remainder of this paper is structured as follows: Section 2 addresses a literature review on the authorship attribution, reviews the text mining models utilized in the state-of-the-art of authorship attribution and also introduces the linear-complexity text classifiers utilized in authorship attribution. Section 3 proposes the intended multi-biometric system as a fast and accurate FDA system by presenting the abovementioned fusion idea. The experiments specifications such as the code complexity, the utilized datasets, and the evaluation metrics are addressed in section 4. Section 5 is devoted to reporting and analyzing the results of the corresponding experiments. Finally, section 6 ends the paper by presenting the concluding remarks.

## 2. Literature Review

In this section, we review the most well-known studies of authorship attribution, which are related to the scope of this paper. About the signature modality, although numerous studies have been conducted in the state-of-the-art [10], considering that the focus of this paper is on the stylome modality, we take advantage of the study of Kudłacik and Porwik [9] as one of the most robust state-of-the-art methods of signature recognition, which has linear-complexity in both of the training and testing phases. Nevertheless, the structure of this section is as follows. In the first subsection, we review the most important studies in authorship attribution. Then, we address the utilized text mining models in authorship attribution. Finally, addressing the linear-complexity text classifiers utilized in authorship attribution ends this section.

### *2.1. Authorship Attribution Studies*

Numerous research studies have been conducted in the state-of-the-art of authorship attribution which are covered and structured by a number of well-known surveys [11][12][13][14][15][16][17][6].

In 2008, Luyckx and Daelemans [18] propose an authorship attribution approach, utilizing memory-based learning, which is specialized for the problems with many authors and limited data for the train, which overcomes SVM and maximum entropy. Also, Iqbal et al. [19] propose a novel method with data mining approach for modeling the stylome of every author as a combination of different frequent features in e-mails. They prove that their method is able to attribute the authors of real-life emails, effectively. In 2010, Layton et al. [20] propose a number of novel preprocessing methods which attribute the authorship, meaningfully superior to the mentioned benchmark in microblog-sized data. Moreover, Raghavan et al. [21] propose a new method for author attribution, by means of probabilistic context-free grammars. For each author, they build a probabilistic context-free grammar and utilize it as a language model for the classification purposes. In 2012, Brenan et al. [22] publish two corpora for adversarial stylometry with 57 unique authors and after theoretically discussing/taxonomizing different methods, prove the high accuracy of four claimed methods for this task. In 2013, Brocardo et al. [23] present a supervised learning technique (as classifier) and n-gram model (as text mining model) for authorship verification in small-sized texts. Their evaluation by means of the Enron email dataset proves a low Equal Error Rate (EER) for 500-character message blocks. In 2014, Sidorov et al. [24] propose the utilization of "syntactic n-gram" (sn-gram) as machine learning features for authorship attribution and prove the very high accuracy, after utilizing sn-grams. In 2015, Amancio [25] proposes the method of the complex network, for upgrading the current statistical methods. By means of fuzzy classification, he proves that the extracted topological properties from texts can complete the standard textual representations. Considering the genericness of his model, his proposed framework can be used to study similar textual applications. In 2015, Segarra et al. [26] propose a method, based on function word adjacency networks. They assume nodes as function words and assumed directed edges from a source function word to a target function word as the likelihood of finding the target in the ordered neighborhood of source. They prove the superior accuracy of their attribution method to the methods that rely on word frequencies. In 2016, Savoy analyzes the authors' distribution and illustrates that it can be modeled by a mixture of two Beta distributions. He approaches this to assign a more accurate probability to the closest author. Then, he proves the overall superiority of his method, by experimenting on the *state of the union* and *federalist papers* datasets. Moreover, Overdorf and Greenstadt [27] address the cross-domain authorship attribution in which the documents have different properties/domains. They utilize blog entries, Twitter feeds, and Reddit comments, as the domains and illustrate that the efficiency of the state-of-the-art authorship attribution methods is not high in cross-domain problems (despite their good accuracy in in-domain problems). Then, they propose specialized methods for the cross-domain problems (both in feature and classification). Also, Peng et al. [28] address the astroturfing problem (appearance of someone in numerous contexts in social media) as a sub-problem of authorship attribution. They utilize binary n-gram text mining model and show how different social network users can be detected as being the same author. In 2017, Shrestha et al. [29] propose attribution of tweets authorship via Convolutional Neural Networks over character n-grams. They propose a strategy for improving the interpretability of models through importance estimation of fragments of input text while classification. Their experiments prove the superiority of their method over the state-of-the-art. Moreover, Markov et al. [30] improve the cross-topic authorship attribution by means of some preprocessing steps on character n-grams as well as a tuning process on features number, focusing on the cross-topic sub-problem. Also, Stamatatos [31] proposes a new method for improving the efficiency of authorship attribution by means of adding a text distortion step, prior to extraction of stylometric measures. His method masks topic-specific information which is unrelated to the authors' personal style. Based on the experiments on closed-set authorship attribution and authorship verification, he demonstrates that his approach enhances the state-of-the-art methods, specifically in the cross-topic conditions (i.e. the condition of topic mismatch in training and testing data). In addition, Posadas-Durán et al. [32] propose a distributed representation for authorship attribution at document-level. Their presented method (at the document level) is trained by distributed vector representations and takes advantage of SVM for the classification purposes. They prove their method to be efficient in six state-of-the-art datasets. In 2018, Grabchak et al. [33] presented a novel approach for one2one authorship verification. They present utilization of an entire profile of lexical richness indices for this purpose. They validate and prove the efficiency of their methodology on several known-author poems. Moreover, Gómez-Adorno et al. [34] suggest training the system by n-gram-based document vectors. For this purpose, they utilize the Paragraph Vector (as a recently proposed) method. The utilized n-grams in their method can be word n-grams, character n-grams, and POS-tag n-grams. Then, they address cross-topic authorship attribution task and validate their method on the well-known Guardian corpus and prove the superiority of the method to its state-of-the-art counterparts.

## 2.2. Text Mining Models in Authorship Attribution

Recognition of authorship in Forensics, as well as other applications (e.g. Stylometry), addresses a behavior-based authentication of identity. For this purpose, a linguistic profile is required. The mentioned linguistic profile is a sufficient number of text data that can be used in order to train a classifier such as Support Vector Machine (SVM), Bayesian Classifier (BC), Adaptive Boosting (AdaBoost) and etc. Correspondingly, the authorship can be determined by means of such classifiers [35]. The Stylometric features can be distinguished in the following main categories. Lexical features (e.g., word n-gram frequency), character features (e.g., character n-gram frequency), syntactic features (e.g., part-of-speech tag frequency), semantic features (e.g., semantic dependency measure), and application-specific features (e.g., specific word frequency, etc.) [24]. A number of studies have shown that the most effective measures are lexical and character features [24]. Recently, Sidorov [24] introduced a new lexical feature named "syntactic n-gram" (sn-gram). The difference between the traditional word n-grams and sn-grams is related to the manner of what elements are considered neighbors. In sn-grams, the neighbors are taken by following syntactic relations in syntactic trees, while n-grams are formed as they appear in texts. In authorship attribution (using SVM), for several profile sizes, sn-grams have been proven to outperform the n-grams of words, POS tags, and characters, when the documents are large (books, novels, etc.). However, for smaller-length documents such as the social-media documents lexical features like n-grams have been proven to have the best performance [36].

## 2.3. Linear-complexity Text Classifiers

In general, computing approach to authorship attribution is divided into literary authorship identification and machine-learning-based text classification [37]. The most common practice of authorship attribution is in supervised learning, in which, the textual documents of authors are modeled by their stylistic features. Then, a classifier is trained by the known textual documents of candidate authors, and at the end, the trained classifier is used to determine the stylistically-closest author to the questioned document [38].

In this section, we address the classification phase of the machine-learning approach in authorship attribution, with a focus on the time complexity. Considering that, in this paper, we aim to propose a multi-biometric system that (in both of the training and testing phases) has linear time-complexity, we review the classifiers with respect to their computational complexity.

Mathematically speaking, letting '$n$' stand for the number of training samples, '$m$' for the number of features of each sample, and '$c$' for the number of classes, we intend to choose a classifier that its time complexity is less than or equal with $O(n \cdot m \cdot c)$. For this purpose, we first enumerate a number of the well-known classifiers, utilized in the state-of-the-art of authorship attribution and then choose the best one for the expected linear-complexity multi-biometric system.

Neural networks are enumerated among the very common classifiers, utilized in authorship attribution systems [39] and can be trained and tested in $O(n \cdot m \cdot c)$, if these three parameters are the only addressed parameters. Naïve Bayes classifiers which utilize the probability distributions that can be estimated in linear time are also considered as linear-complexity classifiers. They include Multinomial Naïve Bayes (MNB) [40] and Poisson Naïve Bayes (PNB) [41] classifiers, whereas some other classifiers such as Weibull Naïve Bayes require at least log-linear complexity for a (reasonably accurate) estimation of the parameters of the utilized probability distribution[2]. K-nearest-neighbors (KNN) classifier [42] is also a linear complexity classifier that depending on the utilized version can perform the classification in $O(n \cdot m)$ (memory-less) or $O(\log(n))$ (memory-full version). Moreover, logistic regression is another classifier utilized in authorship attribution [27] that has linear complexity[3]. Boosting algorithms such as AdaBoost or gradient boosting can also function (be trained/tested) in linear complexity, given that their weak learner algorithm has linear time complexity. Fast (approximated) SVM is also a modification of the support vector machine that can function in linear complexity, while providing accurate results, with a near accuracy to the original version of SVM.

---
[2] Demonstrated in a submitted article of the authors.

There are also some classifiers that cannot be trained and/or tested with linear time complexity with respect to $m, n,$ and $c$ [4]. The original version of SVM is one of those classifiers, considering its requirement to an optimizer such as Sequential Minimal Optimization (SMO) algorithm. Decision trees and correspondingly Random Forest classifiers include the factor $O(n \cdot \log(n))$ [43]. Linear (and Quadratic) Discriminant Analysis (LDA and QDA) classification algorithms are polynomial with respect to $\min(m,n)$ [44][5]. Restricted Boltzmann Machine (RBM) is also another well-known classifier that despite its good accuracy does not function in linear computational complexity.

Table 1 represents a summary of the discussed classifiers in this subsection.

**Table 1.** The list of the discussed classifiers in subsection 2.3 (common classifiers in authorship attribution). It is specified if each classifier has linear time complexity while both of the training and testing phases.

| Linear complexity classifier while the training and testing phases with respect to $n, m$, and $c$ (i.e. $O(n \cdot m \cdot c)$ in both the training and testing phases). | |
|---|---|
| **Yes** | **No** |
| K-nearest neighbors | Support Vector Machines |
| Multinomial and Poisson Naïve Bayes | Decision Tree |
| Fast Support Vector Machine | Random Forest |
| Logistic regression | Linear (and Quadratic) Discriminant Analysis |
| AdaBoost and gradient boost | Weibull Naïve Bayes |
| Artificial Neural Networks | Restricted Boltzmann Machine |

## 3. The proposed multi-biometric system

In this section, we first present the adopted text mining model, followed by the adopted classifier. Then, we propose the mentioned linear-complexity multi-biometric system.

### 3.1. The adopted text mining model

As mentioned in the previous section, choosing the text mining model for authorship attribution depends on the type of the analyzed document. If the document is large (e.g. book or novel), then choosing the sn-gram features is the most effective option for authorship attribution. In contrary, if the document is small (e.g. letters or emails or microblog posts), then lexical features such as n-grams can perform as the best model. Correspondingly, in this paper, we choose the text mining model based on the addressed dataset, as we apply our experiments on different datasets.

### 3.2. The adopted classifier

Considering that, in this paper, we intend to utilize a linear-complexity classifier with respect to $m$, $n$, and $c$, we have to choose the classifier from the first column of Table 1. Although the k-nearest-neighbor classifier is very fast, it is not accurate enough for being a benchmark for authorship attribution experiments [45]. Artificial neural networks, although theoretically can be trained in linear time (with respect to $n, m,$ and $c$), yet in practice, their complexity is dependent on other factors that, meaningfully, affect their complexity. For example, assuming that the utilized artificial neural network has $k$ layers and approximately $d$ nodes in each layer, then its training complexity is polynomial with respect to $d$ [46]. Also, boosting algorithms such as AdaBoos or Gradient Boosting, despite their better accuracy than the utilized weak learner, are not welcomed to be utilized in the proposed multi-biometric system of this paper, because they are two-phase algorithms and (in the best case) include a linear algorithm as "a phase of them," letting alone the complexity of the booster, itself. Moreover, logistic regression (despite its linear complexity) is limited to the two-class classification problem, whereas its multi-class version (i.e. multinomial logistic regression or MaxEnt classifier) is slow for classification problems in which the number of classes are many [47].

However, Naïve Bayes and approximated SVM are, both, linear and also appropriate for our application, as they do not impose any additional complexity burden on the authorship attribution process. Naïve Bayes classifiers function linearly only if their utilized probability distribution can be estimated linearly. Among the utilized probability distributions, (as mentioned) PNB and MNB are linear-complexity and are chosen as the candidates of this paper, beside the approximated SVM. However, Rennie et al. [48] and Ting et al. [49] show that MNB can function as accurate as the other well-known classifiers

---

[3] Logistic regression is a binary classifier and its multi-class version is called multinomial logistic regression or maximum entropy classifier.

[4] Almost all of the classifier algorithms have a trivial linear complexity version. For example, the classifiers that require the solution of an optimization problem, if they choose their optimization algorithm a random one, they can function very fast. In this paper, by linear time complexity, we mean the classifiers that have a standard version, utilized in the state of the art that can function (in both of the training/testing phases) in linear computational complexity.

[5] There is a modification of Linear Discriminant Analysis, called SRDA for reducing its complexity. However, in this paper, we are going to choose the to-be-fused classifier among the well-known classifier.

(e.g. SVM) in the case that its specifications are appropriately set.

Correspondingly, we choose MNB as the main linear-complexity classifier of this study and set the PNB as well as the fast SVM as the benchmarks. Albeit, in practice in the experiments section, we utilize the original SVM for having fairer and more confident comparisons.

### 3.3. The classification process

For each [feature, subject] pair (e.g. ["remember-hour", Peter]) the MNB authorship attribution system counts the feature frequency (e.g. "remember-hour") in the documents of the subject (e.g. Peter) and assigns a probability parameter ($p_{f_i,s_j}$) to that [feature, subject] pair.

$$p_{f_i,s_j} = \frac{N_{f_i,s_j} + \alpha}{N_{s_j} + \alpha \cdot n} \quad \text{Eq. 1}$$

In Eq. 4, $N_{f_i,s_j}$ stands for the number of times that the feature '$f_i$' occurs in the training data that belongs to the subject $s_j$, $\alpha$ stands for the smoothing variable, $N_{s_j}$ stands for the total frequency of all features in the documents of the subject $s_j$, and $n$ stands for the vocabulary size (e.g. the number of total n-grams or sn-grams).

After the training phase, for each [subject], there would be one probability distribution function. Then, having a document ($D_k$), for attribution of its author, we simply compute the above value for different subjects ($s_j$)

$$p(D_k = s_j) = \frac{N_{D_k}!}{N_{f_1,D_k}! N_{f_2,D_k}! \ldots N_{f_n,D_k}!} p_{f_1,s_j}^{N_{f_1,D_k}} p_{f_2,s_j}^{N_{f_2,D_k}} \ldots p_{f_n,s_j}^{N_{f_n,D_k}} \quad \text{Eq. 2}$$

where $p(D_k = s_j)$ represents the probability that the document $D_k$ belongs to the subject $s_j$. Then, the subject $s_j$ with the maximum value of $p(D_k = s_j)$ is assigned as the genuine author of the document $D_k$.

### 3.4. The Multi-biometric FDA system

As mentioned, in this study the fusion of the stylome and the signature modalities is considered as score-level. Having the "stylome match score," (Eq. 2) the other score, required for the score-level fusion is the "signature match score."

As mentioned, the research study of Kudłacik and Porwik [9] is (to the best of our knowledge) the most well-known state-of-the-art signature recognition study that provides high accuracy besides linear-time-complexity (with respect to $n, m,$ and $c$) in both of the training and testing phases. It can consider the fuzziness/ambiguities existing in the signature pixels and provide fuzzy match scores for different [signature, writer] pairs. Correspondingly, we adopt this algorithm for fusion purposes. Utilizing this fuzzy method, the signature match score ($\in [0,1]$) of $(D_k, s_j)$ is considered as $score_{signature}(s_j, D_k)$.

Before addressing the fusion phase, considering that the signature match score is fuzzy, for a better compatibility,

we convert the probabilistic match scores of authorship attribution to a possibilistic/fuzzy version. Based on the probability to possibility relation proposed by Dubois and Prade [48]

$$\pi(s_j) = \sum_{a=1}^{S} \min(p(s_a), p(s_j)). \quad \text{Eq. 3}$$

where $\pi(s_j)$ stands for the possibility of the set $s_j$ and $S$ stands for the total number of sets. We consider subjects as sets and their documents as the members of those sets. Then, based on the defined postulation in [49], we would have

$$\mu_{s_j}(D_k) = \sum_{a=1}^{Subjects} \min\left(\mu_{s_a}(D_k), \mu_{s_j}(D_k)\right). \quad \text{Eq. 4}$$

Utilizing the Eq. 4, we convert the probability match scores of the Eq. 2 to fuzzy match scores. We also do the same probability to possibility conversion for the SVM (the benchmark method) during the proposed fusion experiments.

Now, we have two fuzzy match scores, both of which in [0,1], for stylome and signature compatibility of $(D_k, s_j)$ respectively. Score-level fusion schemes are grouped into three main categories: Density-based, Transformation-based, and Classifier-based [50]. While Density-based schemes require many training samples for genuine/impostor match scores for accurately estimating the density functions, this condition is not available in most of the multi-biometric systems regarding the costs of collecting labeled multi-biometric data; a common circumstance under which Transformation-based schemes are enumerated as viable alternatives [50]. While most of the Transformation-based scheme techniques focus on accurate score normalization for providing compatible scores, in this study, both of the match scores are intrinsically normalized and have a same range ([0,1]) and can be directly fused by standard fusion operators such as "sum the of scores" or its scaled version (average of scores), "product of scores" and etc. [50], among which we use the average of scores standard fusion operator

$$score_{final}(s_j, D_k) = \frac{\mu_{s_j}^{stylome}(D_k) + \mu_{s_j}^{signature}(D_k)}{2}. \quad \text{Eq. 5}$$

Now, obviously, the document $D_k$ will be assigned to a subject with the max score:

$$D_k \to s_l \mid j \in \arg\max_j score_{final}(s_j, D_k), \quad \text{Eq. 6}$$

which ends the proposed multi-biometrics system for being utilized in FDA. Figure 1 provides a perspective to the whole proposed FDA system.

## 4. Experiments Specification

The algorithm is implemented in Python v.3.6.1. The only adopted external libraries are SpaCy[6] and Scikit-learn[7]. SpaCy provides the syntactic parser for sn-gram and Scikit-learn provides MNB (sklearn.naive_bayes.

---
[6] https://spacy.io/

[7] http://scikit-learn.org/

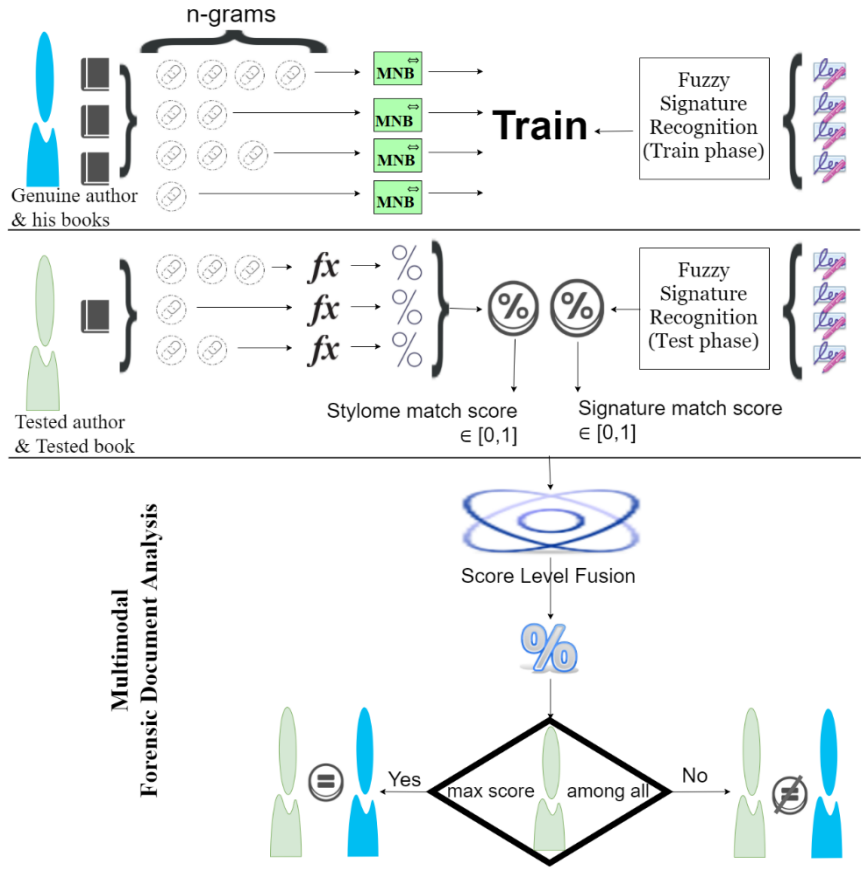

**Figure 1.** A schematic view to the whole multimodal Forensic Document Analysis system. It is proposed in 3 layers / phases: Train, Test, and Fusion.

MultinomialNB) and SVM (sklearn.svm.SVC) classifiers. Moreover, the Poisson classifier [31] is re-implemented in Python.

### 4.1. Code complexity

About the code complexity, given the sn-grams or n-grams, the complexity is linear (as discussed in the previous sections). However, evidently, the complexity of constructing sn-grams should be added to the whole code complexity. Although to the best of our knowledge, the complexity of SpaCy syntactic parser is not mathematically discussed in the state-of-the-art, it is experimentally proven to be the best among other implemented (available) syntactic parsers [51]. We, moreover, should know about the complexity of constructing sn-grams from a parsed document. Although a recursive depth-first traversal algorithm is presented in [24] for construction of sn-grams from syntactic trees, alternatively, we can also utilize its non-recursive version for having linear complexity in, both, time ($O(t + g)$) and space ($O(t)$), when $t$ stands for the number of nodes in the extracted syntactic tree, and $g$ stands for the number of (grammatical) edges. Albeit, the breadth-first search could, also, be utilized for constructing sn-grams, because the order of sn-grams is not important and they are treated as being in a bag. However, accordingly, the total complexity of the Naïve Stylometry experiments is as low as

$$O(c \cdot n \cdot m + c \cdot n \cdot s \cdot (t + g) + SpaCy\_parser) \quad \text{Eq. 7}$$

where $c$ stands for the number of authors, , $n$ for the number of documents of each author, $m$ for the average number of features of each document, $s$ for the average number of sentences of each document and $SpaCy\_parser$ stands for the (low) computational complexity of SpaCy syntactic parser. Please note that, even we can utilize the dependency parser proposed in [52] to make the parser complexity, linear, and therefore the total complexity as

$$O(c \cdot n \cdot m + c \cdot n \cdot s \cdot (t + g) + c \cdot n \cdot s \cdot w) \quad \text{Eq. 8}$$

where $w$ stands for the average number of words of a sentence. For, even, more improvement in time-complexity, GPU programming can be utilized. Then, other parsers such as [53] would be more appropriate. However, this improvement is left to be done in future works of this study.

Please note that the abovementioned analysis is only required for authorship attribution of the large-documents for which we adopt sn-grams. For short-length documents, we utilize n-grams that can be generated in linear time with respect to $s$.

### 4.2. Datasets

In this section, we address the utilized datasets in the experiments of this research and then address the mentioned metrics for evaluation of the proposed system. We first address the experiments which are done on the authorship attribution problem and then address the FDA experiments related to the proposed multi-biometric system. Then, discussing on the evaluating metrics would be the last part of this subsection.

**Authorship Attribution Dataset.** For the authorship attribution phase, we use two different datasets. For large-documents, we utilize Project Gutenberg dataset[8] [54], as a standard dataset, which has also been used in the study of Sidorov et al. on authorship attribution by syntactic bigrams [24]. Sidorov et al. utilize 13 books of three authors and choose eight books for training and five books for testing. Although the mentioned study [24] is one of the most well-known research papers of this field (due to its very efficient proposed text mining model), yet the number of authors that are chosen in [24] is very few (only three authors). To provide more trustable results, in this paper we utilize the entire Gutenberg dataset, including 3,036 English books written by 142 authors. However, choosing the 8-5 train-test proportion bounds our choice from the available 142 authors to those authors who have at least (8+5=) 13 books in the Gutenberg dataset. Applying this filter, 70 authors are removed and correspondingly the adopted dataset includes 72 authors. However, since some of the authors have more than 13 books, to have a comparable style with the 8/5 train/test style utilized in [24], we keep the 13 largest books of each author and remove the remainder. Thus, the adopted classifier has totally 13×72 books, equal with 936.

It is also notable that the experiments related to this dataset are performed by a 13-fold cross-validation, by moving the 8+5 window of train/test proportion, one step by one step.

**Forensic Document Analysis Dataset.** As mentioned in the introduction, the proposed multi-biometric system addresses the FDA problem that is the problem of analyzing the documents written by different subjects with the aim of finding its genuine writer. Considering that the idea of fusing the stylome and the signature biometrics is new and we could not find a structured dataset that has both of the signature and the letters texts, we approach generating a chimeric dataset. Generating chimeric datasets are always an available option for the multi-biometric systems in which there is no correlation between the utilized modalities. Since we believe that there is no correlation between the linguistic style of an author and his signature drawing style, we see this option as the most reasonable one for the evaluating dataset.

Creating this chimeric dataset requires an authorship attribution dataset and one signature recognition dataset. For the stylome modality, we should choose a dataset in which the documents are as short as letters (e.g. a few paragraphs). The CCAT dataset [55] is one of the most well-known short-length authorship attribution datasets. It includes the short-length documents of 50 authors and contains 50×50 documents for the training and 50×50 documents for the testing phase[9].

For the signature modality, we choose the same (SVC) dataset utilized in the mentioned fuzzy signature recognition study [9]. It includes 40 writers each of whom having 20 genuine signatures[10].

For merging these two datasets and creating the final chimeric dataset we set the smaller dataset (SVC) as the basis and by adopting the same approach as what Kudłacik and Porwik [9] adopt, we set the first five signatures of each writer as the training and the remainder 15 signatures as the testing items. Then we pair the CCAT dataset with SVC. We cut the CCAT by choosing its first 40 authors and removing the last 10, and for each author, we cut his documents list by choosing the first five training documents of the training set and removing the last 45 documents, and similarly, choosing the first 15 documents of the testing set and removing the last 35 documents. This finalizes the creation of the utilized chimeric dataset.

### *4.3. Evaluation Metrics*

In this paper, we adopt five evaluating metrics, as follows. The most common and expected metric for both of the authorship attribution and FDA experiments is accuracy that is the proportion of the correctly identified items, divided by the total considered items.

We also use a number of information retrieval metrics. Although the proposed chimeric dataset does not include any imposter item, we can suppose each of the genuine items to be claimed by another writer/author. In other words, for each of the 40×15 test items, we can consider the test item to be claimed by all of the 40 authors/writers (i.e. assuming 40×15×40 claims). Then, we verify the claim if the claimer is the genuine author/writer of the letter or not. It results in 23,400 imposter and 600 genuine claims.

F-score is 2TP/(2TP+FP+FN) when TP stands for true positive, FP for false positive, and FN for false negative detections. F-score is the first chosen retrieval measure in this paper. It is used to evaluate the balance between the precision and recall of the system.

Moreover, we represent the Detection Error Tradeoff (DET) graph which plots the false rejection rate vs. false acceptance. Then, the plot with the smallest integral (area under plot) represents the best one.

In addition, Cumulative Match Characteristic (CMC) curve is the other plot which measures how top-ranked the position of the genuine item in the ranked list of match scores is. The sooner the CMC curve approaches its horizontal asymptote (y=1), the better the ranked match list of the multi-biometric system is.

Also, Match Score Histogram (MSH) is another measure for evaluating how reliable the detection decision of the multi-biometric system is. The farther the distance of the genuine MSH from the imposter MSH is, the more reliable decisions are made by the multi-biometric system.

---

[8] https://web.eecs.umich.edu/~lahiri/gutenberg_dataset.html
[9] https://archive.ics.uci.edu/ml/datasets/Reuter_50_50
[10] Each user also has 20 imposter signatures that we discard, as it is out of the scope of the discussed problem in this paper.

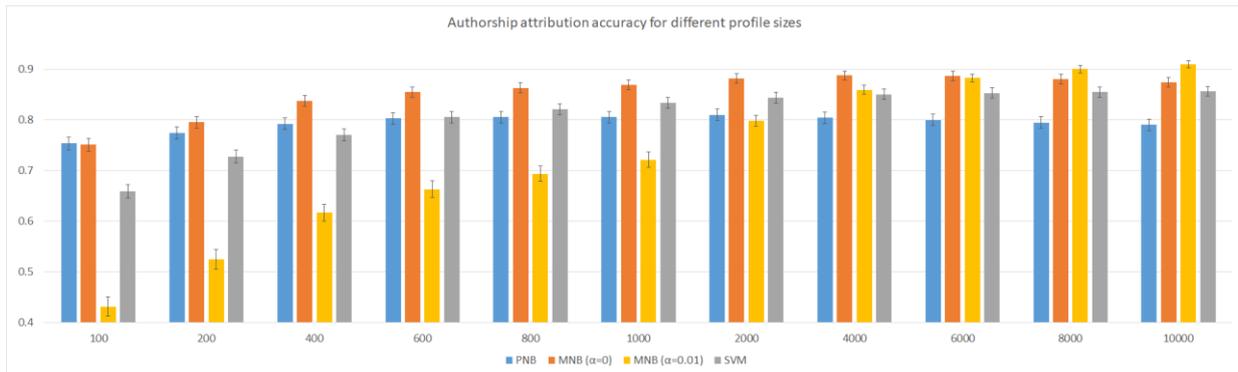

**Figure 2.** The accuracy (and the 95% confidence intervals) of authorship attribution for the best linear-complexity classifiers (i.e. PNB, MNB ($\alpha \in \{0, 0.01\}$), and SVM) and the effect of different profile sizes on the accuracies. MNB with $\alpha = 0$ has the general upper hand and MNB with $\alpha = 0.01$ provides the best accuracy.

# 5. Results and Discussions

This section addresses the authorship attribution and FDA experiments, promised in the previous section. In the first subsection, we address the related experiments to the authorship attribution on the Gutenberg dataset. Then, the second subsection focuses on the FDA experiments on the constructed chimeric dataset, described in section 4.

## 5.1. Authorship attribution

In this subsection, we aim to check the expected superiority of the accuracy of MNB to its counterparts. Moreover, we check the effect of varying the profile size, as well as, varying the sn-gram type on the mentioned expected accuracy.

Fig. 2 illustrates that (as expected) MNB has always the best accuracy, in comparison with SVM and PNB. Moreover, it can be seen that PNB has a promising start and its accuracy competes with the others when the number of features is very few. However, it loses its superiority very soon by the increment of the features number. It may be concluded that for the problems in which, on the one hand, the document size is large, and on the other hand, the speed is very important PNB is recommendable. Nevertheless, for the main addressed problem in this paper (i.e. FDA) this potential conclusion is not of high importance.

Moreover, it can be seen that SVM (albeit with a negative concavity) has an increasing manner in accuracy by the increment of the features number. It can be concluded that the fairest comparisons for the remainder of the experiments would be for the case that the number of features is set to maximum. Thus, we set the features number as 10,000 for the remainder of the experiments.

Finally, it can be seen that the capacity of the accuracy of the MNB with $\alpha = 0$ is saturated when the features number is around 5,000 and it loses a part of its superiority, from then on. However, it is superior enough not to lose its excellence even after 5000. For MNB with $\alpha = 0.01$, it can be seen that despite its low accuracy for the small profile sizes, its accuracy has a fast growth by the increment of the profile size. In other words, when the number of features is large, MNB has a better performance with $\alpha = 0.01$ than $\alpha = 0$. It is because, unlike the cases dealing with small profile sizes, while dealing with large profile sizes, the features-subjects frequency matrix, extracted from the training data is sparse and correspondingly a number of the estimated probability parameters are extremely small, imposing an artificial reductive impact on the total probability. Therefore, avoiding a Laplace smoothing variable forces MNB to lose a part of its potential superiority to its linear-complexity counterpart classifiers.

However, for the small profile sizes, when each feature has meaningfully large occurrences, there is no requirement for such additive smoothing, and reversely, Laplace smoothing can even impose rather a negative effect on the final accuracy.

It is also worthy to note that we test different possible values for the different parameters of the other counterparts and report the best one in Figs. 2 and 3. For example, Scikit-learn SVM has a penalizing parameter which has been initially suggested by Scikit-learn to be set as one of the 0.01, 1, and 100 values. We have tested these three and the best one (i.e. 0.01) is reported. We also do the same for the smoothing variable of PNB. This manual parameter optimization is done for the counterpart classifiers for providing more accurate counterparts, and correspondingly, providing fairer comparisons. In this way, we would have more powerful proofs for the superiority of MNB to its linear-complexity counterparts.

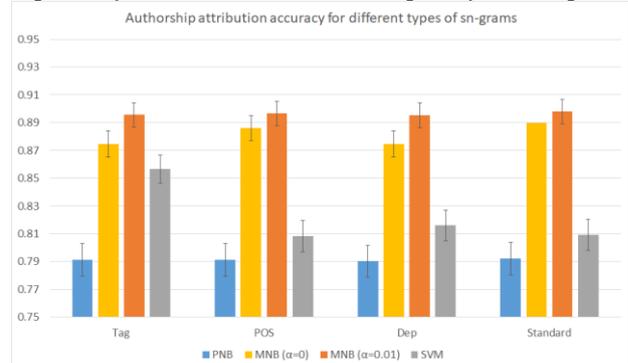

**Figure 3.** The accuracy (and the 95% confidence intervals) of authorship attribution for the best linear-complexity classifiers (i.e. PNB, MNB ($\alpha \in \{0, 0.01\}$), and SVM) and the effect of varying the sn-gram type on the accuracies. MNB has always the upper hand.

Fig. 3 demonstrates that the upper accuracy of MNB over its linear-complexity counterparts do not rely on a specific type of sn-grams and it is preserved by varying them. Moreover, it can be seen that among the four sn-gram types, the best accuracy is related to the standard one. Thus, in the FDA experiments, we set the sn-gram type to be "standard". It is also seen that the superiority of the accuracy of MNB when $\alpha = 0.01$ (to the case with $\alpha = 0$) is not dependent to the sn-gram type and MNB with $\alpha = 0.01$ is always better than MNB with $\alpha = 0$.

## *5.2. Forensic document analysis*

In this subsection, we address the related experiments to FDA by experiments on the constructed chimeric dataset. Before starting the experiments, we check the mentioned expectation about the excellence of the standard n-gram models over the sn-grams when the length of the documents is short. Considering that the utilized authorship attribution dataset for the creation of the proposed chimeric dataset is CCAT which has short-length documents, we test the accuracy of MNB and SVM on this dataset for sn-gram and n-gram text mining models. It is notable that, due to the non-competing performance of PNB for large profile sizes during the experiments of the previous subsection, we discard the PNB in this subsection and let the SVM to be the main (assumed linear-complexity) counterpart of MNB.

Fig. 4 illustrates that (as expected) sn-grams provide, almost always, the least accuracy in comparison with the standard n-grams. Among the standard sn-grams, the accuracy potency of the unigrams is saturated in the middle (approximately the 800 profile size) and decreases from there on. Except for unigrams, the other feature types have an increasing nature by the increment of the profile size. Bigrams, despite their weak accuracy for the small profile sizes, provide very good accuracies by profile size increment. However, utilizing the union of the unigrams set and the bigrams set has a reasonable treatment, by taking the advantages of the unigrams (in small-sized profiles) and bigrams (in large-sized profile). Thus, for the FDA experiments, we take advantage of this type of features (unigrams ∪ bigrams). It can also be seen that MNB (solid lines) almost-always provide more accurate results than SVM (dashed lines), which is another proof for the best performance of MNB over its linear-complexity counterparts.

About the signature recognition experiments, the specifications are defined the same as the specifications of the signature recognition experiments in [9]. However, they [9] conduct the experiments, for different delta-alpha angle values. But, considering that the experiments of the angle value 25 degrees are meaningfully faster than the others, we set delta-alpha to be 25 degrees.

After setting up the specifications of the authorship attribution phase of the fusion experiments, it is the turn to address the fusion experiments, on the constructed chimeric dataset. We start by evaluating the accuracies over the constructed chimeric dataset.

Fig. 5 demonstrates that, although SVM and MNB cannot provide very high accuracies due to the very few available training data (five short-length documents for each author), yet their fusion with the unimodal signature recognition system yields in their improvement. It is notable that the reported accuracies are the average of 15×40 (test docs × authors) prediction. The result of the Student's t-test for meaningfulness of the superiority of the "Sign + SVM" bimodal system over the "Signature" unimodal system is 0.014%, "Sign + MNB" over "Sign + SVM" is 0.037%, and "Sign + MNB" over "Signature" is 0.000036%, all of which pass the Student's t-test with a high confidence.

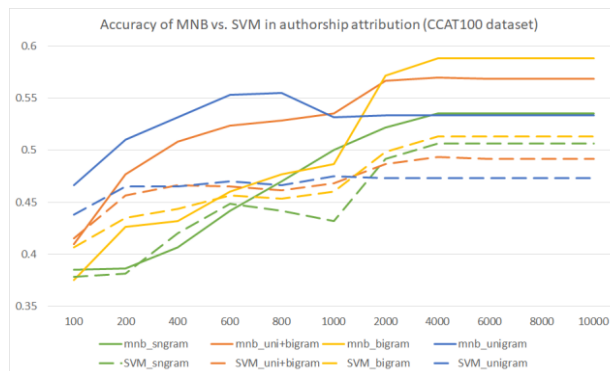

**Figure 4.** Accuracy of authorship attribution yielded by MNB and SVM for different types of features, as well as the effect of profile size on the accuracies. As expected, due to the shortness of the documents length in CCAT, sn-gram do not provide acceptable accuracies. MNB is also always superior to SVM.

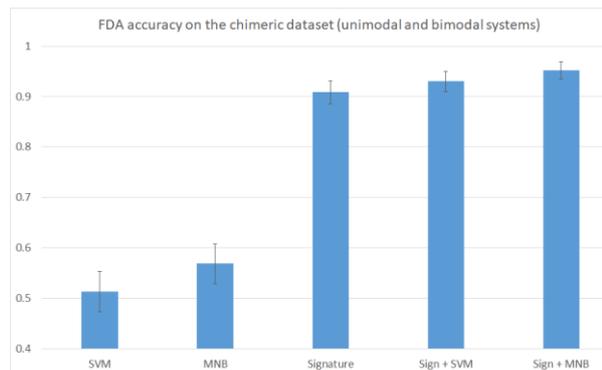

**Figure 5.** The accuracy (and the 95% confidence intervals) of 3 unimodal (SVM, MNB, and Signature) and 2 bimodal (Sign + SVM and Sign + MNB) FDA systems. The bimodal systems meaningfully improve the unimodal signature recognition system.

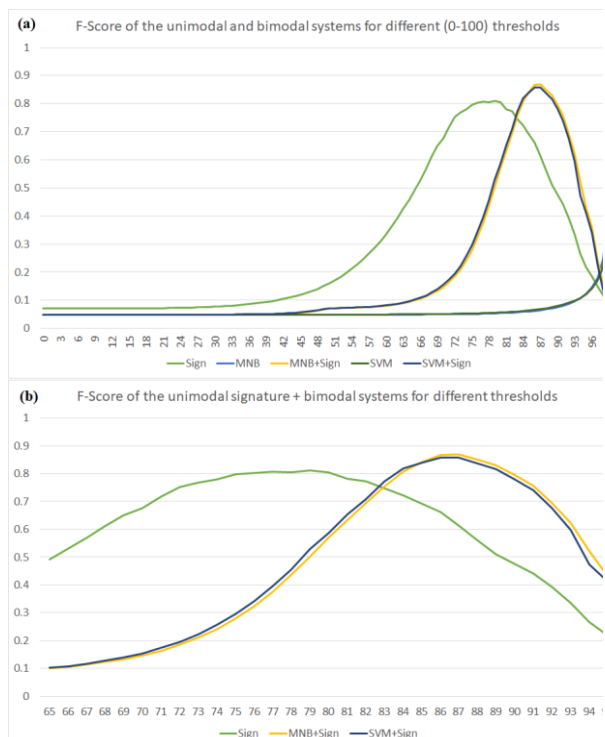

**Figure 6.** The recall of the unimodal and bimodal FDA systems for different threshold values (∈ [0,100]) (part (a)). The bimodal FDA systems meaningfully improve the max recall value of the unimodal "Signature" FDA system (part (b)).

We see that, in the fusion, again, MNB preserves its superiority to SVM and re-proves the correctness of its choice for being fused with the [9] linear-complexity signature recognition algorithm.

Fig. 6 illustrates the recall function of each of the five unimodal and bimodal FDA systems, as a function of the utilized fuzzy membership degree threshold for acceptance/rejection. As it can be seen, although (fuzzy) SVM and (fuzzy) MNB, due to the very few available training data, have unacceptable recall functions alone, the fusion of each of them meaningfully improves the maximum value of the recall function of the corresponding bimodal systems.

Fig. 7 illustrates that, in addition to its improving effect on accuracy and the F-score, the proposed "MNB + Sign" bimodal FDA system, also, improves the CMC curve treatment of unimodal systems. In other words, the convergence rate of the "MNB + Sign" FDA system to its horizontal asymptote is more than the convergence rate of the "Signature" unimodal system. Moreover, it can be seen that, still, MNB has a better performance in comparison with SVM. In other words, the convergence rate of "MNB + Sign" to its horizontal asymptote is even more than the convergence rate of "SVM + Sign," which is a double proof for the superiority of MNB over its linear-complexity counterparts.

Fig. 8 illustrates that the fusion idea alleviates both of the false positive and false negative error rates, belonging to the "Signature" unimodal FDA system, at the same time. As it can be seen, the "MNB + Sign" and "SVM + Sign" DET curves are both meaningfully lower than the DET curve of the "Signature" unimodal FDA system. Again, the "MNB + Sign" is slightly lower in comparison with the "SVM + Sign" bimodal FDA systems which is a double proof for the superiority of SVM over its linear-complexity counterparts.

Fig. 9 illustrates that the fusion idea, even, improves the MSHs. As it can be seen, although MNB and SVM unimodal FDA systems have very low-quality MSHs (having large common genuine/imposter area), the fusion of them with the "Signature" unimodal FDA system provides the best MSH (having the minimum common genuine/imposter area). Also, by comparing the "MNB + Sign" with "SVM + Sign," this fact can be seen that both of the genuine and imposter areas are "slightly" tighter in "MNB + Sign" and therefore it has a slightly-better treatment, the same improvement as seen in the previous plots as well. There is also an interesting fact about the "Signature" unimodal FDA system; one-third of the imposter signatures receive zero match score. This fact shows the strength of the utilized signature method [9], despite its linear-complexity.

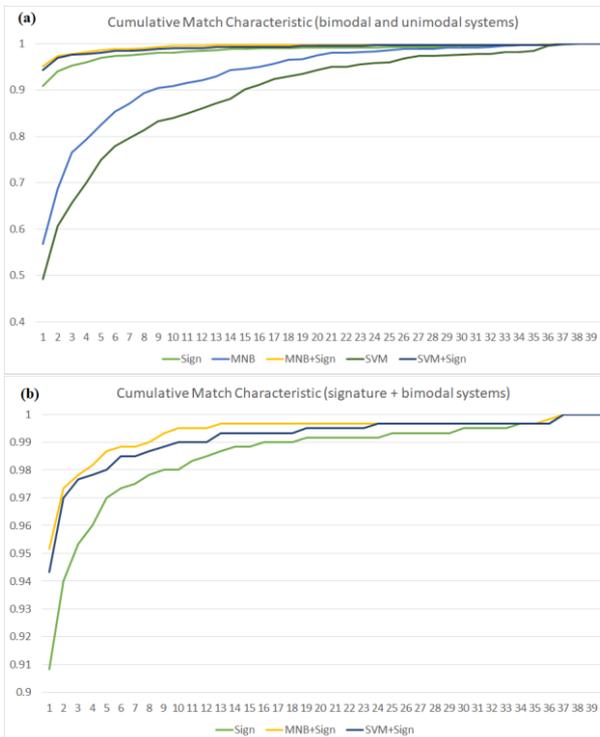

**Figure 7.** The cumulative match score curve related to the unimodal and bimodal FDA systems (part (a)). The bimodal systems meaningfully improve the CMC curve of the "Signature" unimodal FDA (part (b)). As expected, the more improvement belongs to the "MNB + Sign" bimodal FDA system.

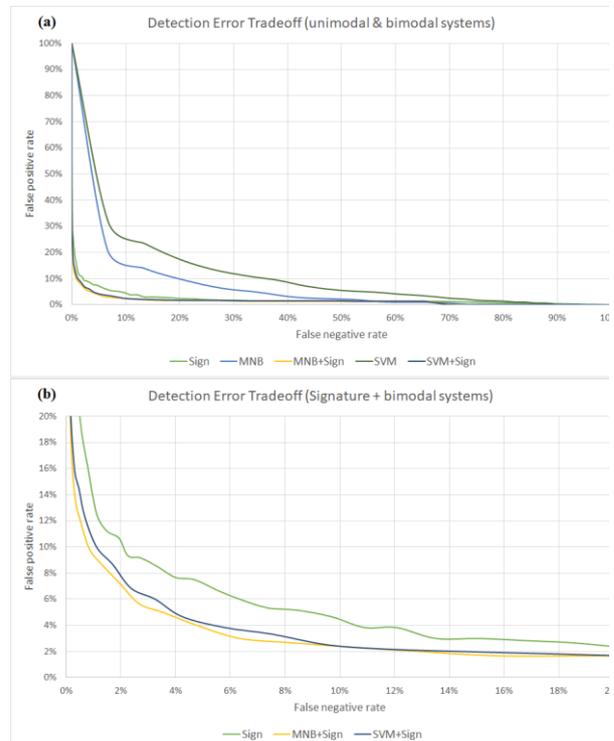

**Figure 8.** The detection error trade-off related to the unimodal and bimodal FDA systems (part (a)). The bimodal systems meaningfully improve the DET curve of the "Signature" unimodal FDA (part (b)). As expected, the more improvement belongs to the "MNB + Sign" bimodal FDA system.

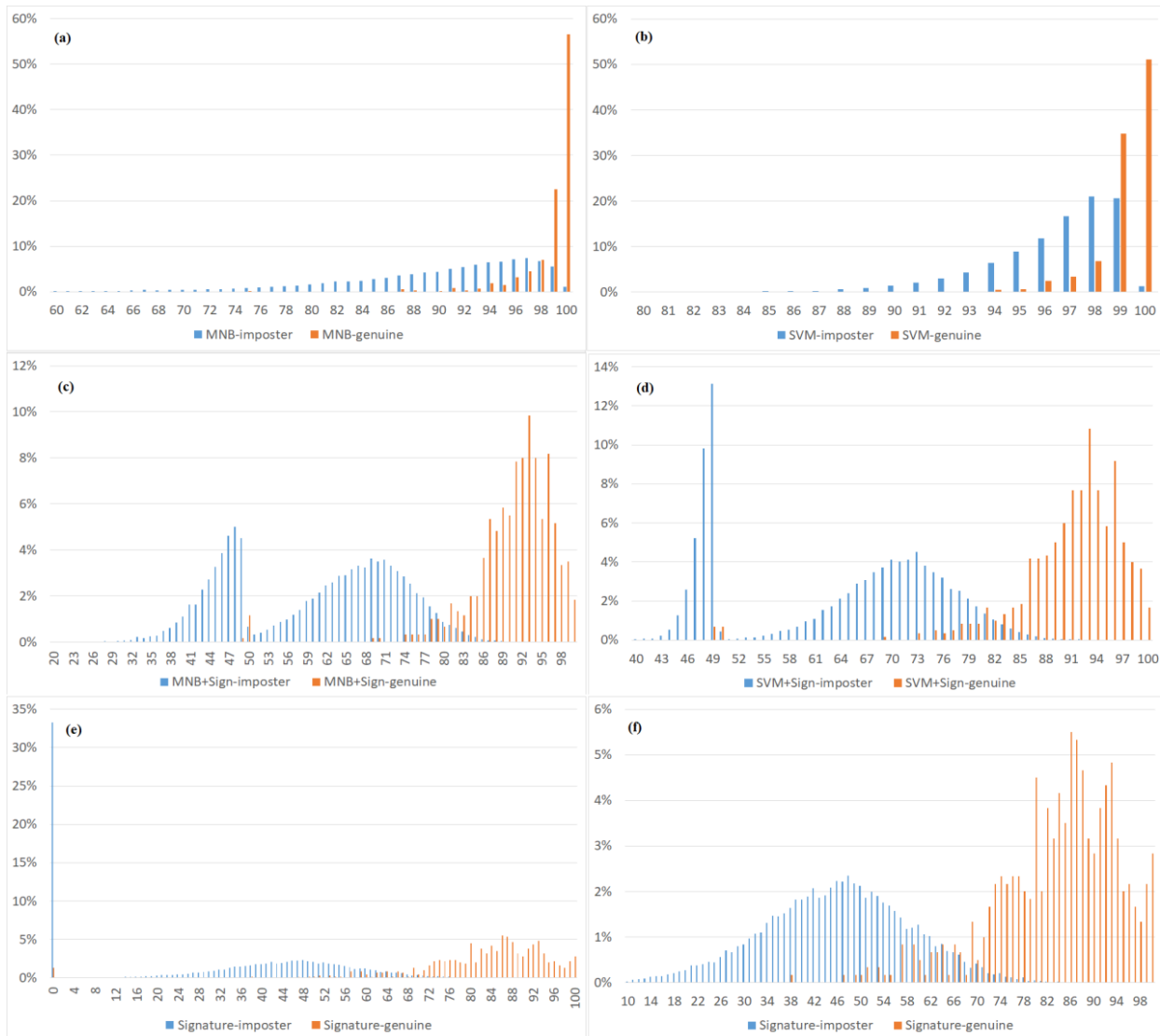

**Figure 9.** Match score histograms, related to (a) MNB, (b) SVM, (c) MNB + Sign, (d) SVM + Sign, and (e,f) Signature. Unimodal authorship attribution systems (i.e. (a) and (b)) have the worst treatment (common imposter/genuine area). Then, "Signature" has a better treatment; and the best treatment (i.e. the least common genuine/imposter area) belongs to "MNB + Sign" and "SVM + Sign) bimodal FDA systems.

# 6. Conclusion and Future Works

In this paper, propose a very precise bimodal Forensic Document Analysis (FDA) system by fusion of authorship attribution and signature recognition, for the first time in the FDA state-of-the-art. The proposed multi-biometric system functions in linear time-complexity for both of the training and the testing phase. We use a well-known linear-complexity and robust state-of-the-art signature identification method for the signature modality and introduce the Multinomial Naïve Bayes (MNB) as the best authorship attribution classifier that can be trained and tested in linear time. In the proposed system, the fuzzified version of the MNB match score is fused (by the average-of-scores fusion operator) with the fuzzy signature match score (both of which membership degrees, falling in [0,1]). The experiments prove the high performance of the proposed system, by evaluating different metrics such as accuracy, F-score, Detection Error Trade-off (DET), Cumulative Match Characteristics (CMC) curve, and Match Score histograms. Due to its linear-complexity, not only the proposed bimodal FDA system is recommended for the FDA problems in which the number of suspicious writers is many, but also it is recommended for real-time FDA applications.

# 7. Acknowledgments

This research has partly been supported by AGAUR research grant 2013 DI 012, IDENTITY– n.690907, Wise Intelligent Agents (AHD) research grant 13960001-01, 139700002, the IN2014-2576-7-R (QWAVES) Nuevos métodos de automatización de la búsqueda social basados en waves de preguntas, and ANSwER: ANálisis de SEntimiento y segmentación de las Redes sociales para la generación de leads y el análisis de complementariedad de marcas, RTC-2015-4303-7; and the grup de recerca consolidat CSI-ref. 2014 SGR 1469.

*the 18th International Conference on Computational Linguistics and Intelligent Text Processing.*, 2017, no. Cic.